\pdfoutput=1

\documentclass[10pt]{article}

\usepackage[preprint]{rlj}

\usepackage{amssymb}
\usepackage{mathtools}
\usepackage{mathrsfs}
\usepackage{graphicx}
\usepackage{subcaption}
\usepackage[space]{grffile}
\usepackage{url}

\title{Operator-on-$F$ complements value-equivalence: a planning-time diagnostic for latent world models}

\setrunningtitle{Operator-on-$F$ complements value-equivalence}

\author{Donna Vakalis\textsuperscript{1,2}}

\emails{donna.vakalis@olympian.org}

\affiliations{
$^{1}$\textbf{Mila -- Quebec AI Institute} \\
$^{2}$\textbf{Universit\'e de Montr\'eal}
}

\contribution{
    A complementary world-model evaluation metric, operator-on-$F$, computed
    natively on TD-MPC2 and LeWM continuous latents at the model's
    planning-time horizon. On a released TD-MPC2 size sweep over
    \texttt{cheetah-run}, it flags a planning-relevant failure mode that
    conventional (unnormalized) reward-prediction and Bellman-residual checks miss.
    }
    {
    Diagnostic only; complementary to value-equivalence rather than replacing it.
    }

\contribution{
    A finding on the released TD-MPC2 mt80 sweep: across five
    sizes ($1$M--$317$M), $317$M has the highest operator-on-$F$ error
    ($2.62$, an order of magnitude above the $0.28$--$0.36$ cluster) and a return
    collapse to $0.9$, while reward-prediction error stays within
    the same small $[0.028, 0.091]$ range across all five. Rank
    correlation between operator error and return loss is $-0.90$
    (anchor-bootstrap $95\%$ CI $[-0.90, -0.70]$; leave-one-out drop of any
    single size leaves Spearman at $-0.80$ or stronger).
    }
    {
    $n=5$ sizes is inherently small; the qualitative observation at $317$M (operator error an order of magnitude above the rest, return collapse) is the most robust element of this finding.
    }

\contribution{
    A re-ranking of the same TD-MPC2 size sweep on real models: full-$F$
    operator error disagrees with the Bellman residual at Spearman $+0.30$
    (anchor-bootstrap $95\%$ CI $[+0.30, +0.60]$), contrasted with the
    near-identity ($+1.00$) between the unnormalized value-only error and
    the Bellman residual.
    }
    {
    Single seed per size; CIs bootstrap over anchors only.
    }

\contribution{
    A brief cross-architecture sanity check on DMC cheetah-run: operator-on-$F$ reports
    informative differences between pure-SSL LeWM (no anchor, $3$ seeds) and
    a task-anchored single-task TD-MPC2 with disjoint $95\%$ CIs that are
    preserved under a $1$-hidden-layer MLP-probe ablation.
    }
    {
    Single environment; two-point architecture spectrum (no reconstruction-anchored midpoint).
    }


\summary{
World-model evaluation in model-based reinforcement learning typically asks
whether the learned model predicts reward and value well. This can be silent
at planning-relevant errors in the model's latent rollouts. We introduce operator-on-$F$, a complementary
diagnostic that compares a model's
$k$-step latent pushforward to the environment's on a chosen observable
subset $F$, computed natively on the model's own predictor at its
planning-time horizon. On the released TD-MPC2 mt80 size sweep over
\texttt{cheetah-run}, reward-prediction error stays within
$[0.028, 0.091]$ across all five sizes — only $\sim 3\times$ variation, so
an unnormalized reward-fit metric has narrow resolution to distinguish them — while
operator-on-$F$ error spans $0.28$ to $2.62$ and tracks the
executed-return loss (rank correlation $-0.90$, anchor-bootstrap $95\%$ CI
$[-0.90, -0.70]$). At $317$M the operator error is an order of
magnitude above the $0.28$--$0.36$ cluster ($2.62$), and the planning return collapses to
$0.9$; reward-prediction error at $317$M ($0.091$) stays within the same small range as the rest of the sweep. The
operator also returns informative, architecture-discriminating estimates in
a cross-architecture comparison between task-anchored TD-MPC2 and a pure-SSL
latent world model (LeWM). We position operator-on-$F$ as
complementary to value-equivalence, not as a replacement: report both rather
than treat either as a complete picture.
}

\begin{document}

\maketitle

\begin{abstract}
World-model evaluation for model-based reinforcement learning typically asks
whether the learned model predicts reward and value well, which can leave
planning-relevant errors in the model's latent rollouts unmeasured. We
introduce a complementary diagnostic, operator-on-$F$, that compares a
model's $k$-step latent pushforward to the environment's on an observable
subset $F$, using the model's own predictor. On a TD-MPC2 size sweep over
\texttt{cheetah-run}, reward-prediction error stays within
$[0.028, 0.091]$ for every model size — only $\sim 3\times$ variation —
so an unnormalized reward-fit check has narrow resolution to distinguish
them; the (unnormalized) Bellman residual and reward error themselves have
weak relationships with return (Spearman $-0.10$ and $-0.30$). Operator error spans $0.28$ to $2.62$ over the same sizes. At $317$M the operator error is
$2.62$ — an order of magnitude above the $0.28$--$0.36$ cluster — and the planning
return collapses to $0.9$, while reward-prediction error ($0.091$) is the
highest of the five but stays within the same small $[0.028, 0.091]$ range
as the rest of the sweep. The rank correlation between operator error and
return loss is $-0.90$ (anchor-bootstrap $95\%$ CI $[-0.90, -0.70]$ at
$n=5$ sizes; leave-one-out removal of any single size leaves it at $-0.80$
or stronger). The operator also returns informative,
architecture-discriminating estimates in a cross-architecture comparison
between TD-MPC2 and a pure-SSL latent world model. The operator diagnostic complements value-equivalence rather than
replacing it.

\end{abstract}

\section{Introduction}
\label{sec:intro}

World-model evaluation in model-based reinforcement learning often treats two
questions as equivalent: does the learned model predict reward and value well,
and does it model the task's dynamics well? Value-equivalence
\citep{grimm2020value,grimm2021proper,silver2017predictron} argues that for
planning, what a model gets right about reward and value is what matters. When
the answers to the two questions agree, training a model to predict reward is
enough. When they disagree, a reward-prediction check can be silent at the failure
mode that matters for planning.

This paper proposes a complementary diagnostic, operator-on-$F$, that
compares the model's $k$-step latent rollout against the environment's at the
level of an observable subset $F$ rather than at the reward head. We then
report results on released TD-MPC2 \citep{hansen2024tdmpc2} checkpoints in
which the two metrics disagree at the largest scale of the
released mt80 sweep: at $317$M the operator-on-$F$ error is an order of
magnitude above the $0.28$--$0.36$ cluster ($2.62$ vs.\ next-highest $0.91$, at $1$M)
while reward-prediction error ($0.091$) stays within the same small
$[0.028, 0.091]$ range as the rest of the sweep, and the planning return
collapses to $0.9$. We also report the same diagnostic on a pure-SSL
world model trained from scratch (LeWM, \citet{maes2026lewm}) used as a
cross-architecture comparison; both of these are single-environment
observations, not a general claim.

Our contributions are: (i) the operator-on-$F$ definition and a probe-based
estimator that runs natively on a model's own latent predictor at its
planning-time horizon; (ii) on the released TD-MPC2 mt80 size sweep,
reward-prediction error stays within the narrow $[0.028, 0.091]$ band
across all five sizes ($\sim 3\times$ variation) while operator error spans
$0.28$ to $2.62$, with the $317$M model showing the highest operator
error (an order of magnitude above the $0.28$--$0.36$ cluster) and a return
collapse to $0.9$ (rank correlation between operator error and return
$-0.90$, anchor-bootstrap $95\%$ CI $[-0.90, -0.70]$, with leave-one-out
drop of any single size leaving Spearman at $-0.80$ or stronger); (iii) a re-ranking of the same five models that disagrees with
the Bellman residual (Spearman $+0.30$, anchor-bootstrap $95\%$ CI
$[+0.30, +0.60]$), contrasted with the per-size near-identity between the
unnormalized value-only error and the Bellman residual (Spearman $+1.00$);
and (iv) a cross-architecture sanity check between single-task TD-MPC2 and
pure-SSL LeWM on \texttt{cheetah-run} with disjoint $95\%$ CIs that are
preserved under a $1$-hidden-layer MLP-probe ablation.

\section{Background and method}
\label{sec:method}

\subsection{Operator-on-\texorpdfstring{$F$}{F}}
\label{sec:c7}

Let $s_t$ be a state, $a_t$ an action, and $z_t = \mathrm{enc}(s_t)$ a
model's latent encoding. We write the model's $k$-step rollout as
$\hat{z}_{t+k} = O_k(z_t,\, a_{t:t+k-1})$ and the environment's encoded
next-state as $z_{t+k}' = \mathrm{enc}(s_{t+k})$. For TD-MPC2
\citep{hansen2024tdmpc2}, $O_k$ is its latent dynamics rolled $k$ times
open-loop; for LeWM \citep{maes2026lewm}, $O_k$ is its native predictor at a
$k$-step frameskip. In both cases $O_k$ is the model's own learned map and is
nonlinear; linearity enters the protocol only on the readout side, through
the probe and the full-$F$ basis below. A probe $\phi_F : Z \to F$ maps latents to a chosen
observable subset $F$ \citep{alain2017probes}, so a fair cross-model
comparison can happen on a surface both encoders share.

For each functional $\phi \in F$ and each anchor
$(s_t,\, a_{t:t+k-1},\, s_{t+k})$, we compute the per-anchor operator error
$|\phi(\hat{z}_{t+k}) - \phi(z_{t+k}')|$ normalized by the across-anchor
spread of the true pushforward, and aggregate by RMS over $(\phi, \text{anchor})$.
The \emph{value slice} sets $F = \{r, V\}$ using the model's own reward and
value heads. The \emph{full-$F$} aggregate adds a held-out per-anchor PCA
basis on the encoded next-state geometry (top-$16$ directions, whitened by
their singular values); the basis is fit on a held-out half of the anchors
and evaluated on the other half so it cannot adapt to operator error.
Appendix~\ref{app:c7} gives the full expression.

\subsection{Epistemic class and probe}
\label{sec:epistemic}

We report all operator-on-$F$ values as \emph{native}: $O_k$ is each model's
own predictor at the same $k$ a planner would use at deployment, and the
comparison is to encoded ground-truth next-states. We never substitute a
synthetic predictor. LeWM has no reward or value head, so for the
cross-architecture comparison we report the operator on a shared
observation-level $F$ (cheetah-run observation plus reward) probed
identically on both models.

The probe is a ridge regression with feature standardization and $\lambda$
chosen by grid search on a held-out sub-split, reported with per-observable
$R^2$ so probe quality is visible. Unregularized probes degrade sharply on
high-dimensional latents (in a pilot, unregularized ridge on the TD-MPC2
mt80-$19$M $768$-dim latent gave strongly negative $R^2$ on some
observables), an instance of the
capacity-vs-information confounds documented in the probing-classifier
literature \citep{hewitt-liang-2019-designing}, which makes unregularized probes
an unfair surface for cross-model comparison. Appendix~\ref{app:probe}
gives the protocol in full, including the commonly-well-probed restriction
used as the probe-asymmetry control in Section~\ref{sec:results-t23}.

\section{Experimental setup}
\label{sec:setup}

We evaluate three sources of world models on the DMC \texttt{cheetah-run}
task \citep{tassa2018dmc}: the released TD-MPC2 \citep{hansen2024tdmpc2}
mt80 multitask checkpoints at five parameter counts ($1$M, $5$M, $19$M,
$48$M, $317$M), used as the size-sweep axis; the released TD-MPC2
single-task cheetah-run checkpoint, used as the single-task
control; and three LeWM (\textit{LeWorldModel}, \citet{maes2026lewm}) seeds
trained from pixels for $100$ epochs each with the SIGReg regularizer of
\citet{balestriero2025lejepa} active. LeWM has no reward or value head, so
the cross-architecture comparison reports the operator on observation-level $F$ only.

Operator error is measured at a matched horizon of $5$ collect-steps. LeWM
predicts the jump in one forward pass (frameskip $5$); TD-MPC2 rolls its
latent dynamics $5\times$ open-loop under the recorded actions. The
cross-architecture comparison (Section~\ref{sec:results-t23}) pools $1552$
evaluation anchors over $16$ held-out episodes; the TD-MPC2 size-sweep
(Sections~\ref{sec:results-t25} and \ref{sec:results-t22}) uses the $1500$
pooled anchors from the size-sweep rollouts. Configurations, analysis
scripts, and figure scripts to reproduce the operator-error,
metric-disagreement, and cross-architecture results reported in
Section~\ref{sec:results} are available
at \url{https://github.com/DonnaVakalis/operator-on-F}.

\section{Results}
\label{sec:results}

\subsection{Value-equivalence can be silent when the operator is wrong}
\label{sec:results-t25}

\begin{figure}[tb]
    \centering
    \includegraphics[width=\linewidth]{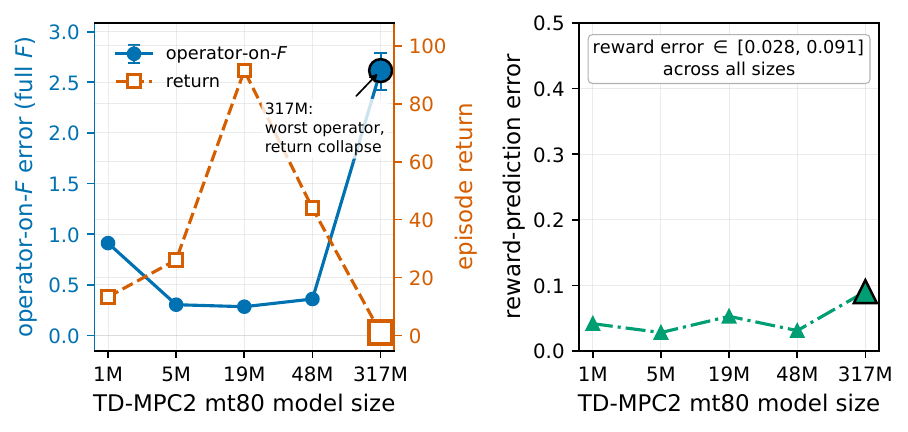}
    \caption{TD-MPC2 mt80 size sweep on \texttt{cheetah-run}. Left:
        full-$F$ operator-on-$F$ error and executed return as a function of
        model size. Right: reward-prediction error on the same axis.
        Reward-prediction error stays within $[0.028, 0.091]$ across all
        five sizes ($\sim 3\times$ variation), while operator error tracks
        the return loss (Spearman $-0.90$, anchor-bootstrap $95\%$ CI
        $[-0.90, -0.70]$; leave-one-out $\in [-1.00, -0.80]$). The $317$M
        model has the highest operator
        error and a return collapse to $0.9$; reward-prediction error at
        $317$M ($0.091$) is the highest of the five sizes but stays
        within the same small range as the rest of the sweep.}
    \label{fig:scissors}
\end{figure}

On the released TD-MPC2 mt80 sweep over five parameter counts
(Figure~\ref{fig:scissors}), reward-prediction error stays within
$[0.028, 0.091]$ at every size — only $\sim 3\times$ variation, narrow
resolution for an unnormalized reward-fit check to distinguish them — while the
full-$F$ operator-on-$F$ error spans $0.28$ to $2.62$. The $317$M checkpoint is the clearest case: operator
error is $2.62$, an order of magnitude above the $0.28$--$0.36$ cluster
(next-highest $0.91$, at $1$M), and the planning return collapses to $0.9$,
while reward-prediction error ($0.091$) remains within the same small
$[0.028, 0.091]$ range as the rest of the sweep. Neither value-equivalence
proxy, as conventionally (unnormalized) reported, orders the models by
return: Spearman $-0.10$ for the Bellman residual and $-0.30$ for
reward-prediction error (normalizing each channel by its target spread, as
operator-on-$F$ does, recovers the return ordering; Appendix~\ref{app:t25}).
Per-size operator-on-$F$
values have tight anchor-bootstrap CIs (Appendix~\ref{app:t25}): the
across-size rank correlation with return is $-0.90$ (anchor-bootstrap
$95\%$ CI $[-0.90, -0.70]$), and leave-one-out removal of any single
size leaves the Spearman at $-0.80$ or stronger, so the rank order is
stable to single-point outliers, though $n=5$ is inherently small.
Because reward-prediction error has limited variance across the sweep,
a partial correlation controlling for it reproduces the marginal value
($-0.89$); we report this as consistency, not an independent control.

\subsection{Operator-on-\texorpdfstring{$F$}{F} disagrees with value-equivalence proxies}
\label{sec:results-t22}

On the same five TD-MPC2 sizes, the full-$F$ operator-on-$F$ rank-disagrees
with the Bellman residual: Spearman $+0.30$ (anchor-bootstrap $95\%$ CI
$[+0.30, +0.60]$), and $+0.10$ ($[+0.10, +0.30]$) against the kernel-divergence
operator. The paper's value slice $F = \{r, V\}$ (normalized per-functional
aggregate) also disagrees with the Bellman residual at Spearman $+0.10$
($[+0.10, +0.20]$); it is the held-out PCA basis that lets full-$F$ pick up
return-tracking structure neither metric sees.

To anchor the comparison: a naive value-equivalence proxy, the per-anchor
unnormalized $|V(\hat{z}) - V(z')|$ averaged across anchors, is numerically
near-identical to the Bellman residual on this sweep
(per size: $2.66 / 1.33 / 2.80 / 2.70 / 6.78$ vs.\ Bellman residual
$2.65 / 1.33 / 2.78 / 2.67 / 6.73$; Spearman $+1.00$). This is the
$\gamma$-scaled equivalence that TD-MPC2's small reward-prediction error
($\mathrm{error} \in [0.028, 0.091]$, nearly two orders of magnitude below the
value-error component) produces between the two quantities.
The normalized value slice and the full-$F$ operator both depart from this
equivalence; the full-$F$ operator is the one whose ranking tracks return
loss (Section~\ref{sec:results-t25}).

\subsection{Cross-architecture sanity check}
\label{sec:results-t23}

To check whether the diagnostic returns informative estimates across
architectures, we compare pure-SSL LeWM ($3$ seeds, no anchor) against the
released single-task TD-MPC2 on \texttt{cheetah-run}, both at the matched
$5$-step horizon and on a shared observation-level $F$ probed identically. LeWM's
operator error on shared $F$ across seeds is $0.384 \pm 0.009$; single-task
TD-MPC2's is $0.840$ ($95\%$ CI $[0.79, 0.89]$); both beat the persistence
baseline ($1.33$ for LeWM, $1.40$ for TD-MPC2), and the $95\%$ CIs are disjoint
(Figure~\ref{fig:crossarch} in Appendix~\ref{app:t23}). A re-run with a
$1$-hidden-layer MLP probe (matched protocol, $\alpha$ grid-searched on
the same sub-split) preserves the ordering with disjoint CIs (LeWM
$0.494 \pm 0.006$, TD-MPC2 $0.997$, ratio $0.495 \pm 0.006$;
Appendix~\ref{app:probe-mlp}), so the gap is not a ridge-probe-capacity
artifact. Probe-asymmetry and prediction-mode controls (Appendix~\ref{app:t23})
also preserve the gap. We do
not draw a broader claim about which family of model the operator metric
favors from a single environment with a two-point architecture spectrum.

\section{Limitations and discussion}
\label{sec:limits}

The cross-architecture comparison is on a single environment
(\texttt{cheetah-run}), and the architecture spectrum is two points
(TD-MPC2 and LeWM) with no reconstruction-anchored midpoint such as
DreamerV3 \citep{hafner2023dreamerv3}; the TD-MPC2 size-sweep correlation
is $n=5$ and reported as correlational. The $317$M return collapse that
anchors that correlation is specific to \texttt{cheetah-run}: on four
sibling mt80 tasks the same $317$M checkpoint plans competently (walker-walk
and reacher-easy return $\sim 800$--$930$) with correspondingly low operator
error, so it is one operating point of a broader operator--return
association (pooled partial Spearman $-0.49$ across the $(\text{task},
\text{size})$ cells) rather than a globally degenerate checkpoint. The probe
is a confound, since a
shared $F$ across different latent geometries must be read out through a probe; we report
probe $R^2$ with each operator-on-$F$, cross-check on the commonly-well-probed
restriction (Appendix~\ref{app:t23}), and verify that a $1$-hidden-layer
MLP-probe variant preserves the cross-architecture ordering with disjoint
CIs (Appendix~\ref{app:probe-mlp}).

Operator-on-$F$ is a pushforward comparison on a chosen observable subset,
in the operator-theoretic lineage of Koopman theory and EDMD
\citep{mezic2005,williams2015edmd,mezic2021}. $O_k$ is nonlinear
(Section~\ref{sec:c7}), so the comparison is not a linear-operator
approximation. In the representation language of \citet{mezic2021},
$\mathrm{enc}$ plays the role of the vector of observables and $O_k$ the role
of the map in the representation eigenproblem---a finite-dimensional
representation that here is nonlinear and reduced, and, since actions enter,
of the control form \citep{korda2018mpc}; our $F$ names the observable set
read out through the probe, not that map. Operator-on-$F$ then estimates, per
anchor, what \citet{mezic2021} calls the accuracy of the representation,
measured on the observables in $F$. \citet{lyu2023latent} train a
task-anchored Koopman latent for control and \citet{ruizmorales2025}
examine Koopman invariants inside JEPA-style models. We use operator-on-$F$ as a
diagnostic on existing models, not as a training objective, and we make no
spectral claims.
We do not claim that value-equivalence
\citep{grimm2020value,grimm2021proper} is wrong, only that on this sweep
and this environment a reward- or Bellman-residual check, as conventionally
reported, is silent at the failure mode the operator metric detects; we
recommend reporting both rather than one in place of the other.

\subsubsection*{Acknowledgments}
We gratefully acknowledge financial support from the Canada CIFAR AI Chairs
program, and compute resources from Mila (mila.quebec).

\bibliography{main}

\begin{thebibliography}{16}
\providecommand{\natexlab}[1]{#1}
\providecommand{\url}[1]{\texttt{#1}}
\expandafter\ifx\csname urlstyle\endcsname\relax
  \providecommand{\doi}[1]{DOI: #1}\else
  \providecommand{\doi}{DOI: \begingroup \urlstyle{rm}\Url}\fi

\bibitem[Alain \& Bengio(2017)Alain and Bengio]{alain2017probes}
Guillaume Alain and Yoshua Bengio.
\newblock Understanding intermediate layers using linear classifier probes.
\newblock In \emph{International Conference on Learning Representations
  Workshop ({ICLR Workshop})}, 2017.

\bibitem[Balestriero \& LeCun(2025)Balestriero and
  LeCun]{balestriero2025lejepa}
Randall Balestriero and Yann LeCun.
\newblock {LeJEPA}: Provable and scalable self-supervised learning without the
  heuristics, 2025.

\bibitem[Grimm et~al.(2020)Grimm, Barreto, Singh, and Silver]{grimm2020value}
Christopher Grimm, Andr\'e Barreto, Satinder Singh, and David Silver.
\newblock The value equivalence principle for model-based reinforcement
  learning.
\newblock In \emph{Advances in Neural Information Processing Systems}, 2020.

\bibitem[Grimm et~al.(2021)Grimm, Barreto, Farquhar, Silver, and
  Singh]{grimm2021proper}
Christopher Grimm, Andr\'e Barreto, Gregory Farquhar, David Silver, and
  Satinder Singh.
\newblock Proper value equivalence.
\newblock In \emph{Advances in Neural Information Processing Systems}, 2021.

\bibitem[Hafner et~al.(2023)Hafner, Pasukonis, Ba, and
  Lillicrap]{hafner2023dreamerv3}
Danijar Hafner, Jurgis Pasukonis, Jimmy Ba, and Timothy Lillicrap.
\newblock Mastering diverse domains through world models.
\newblock \emph{arXiv preprint arXiv:2301.04104}, 2023.

\bibitem[Hansen et~al.(2024)Hansen, Su, and Wang]{hansen2024tdmpc2}
Nicklas Hansen, Hao Su, and Xiaolong Wang.
\newblock {TD-MPC2}: Scalable, robust world models for continuous control.
\newblock In \emph{International Conference on Learning Representations}, 2024.

\bibitem[Hewitt \& Liang(2019)Hewitt and Liang]{hewitt-liang-2019-designing}
John Hewitt and Percy Liang.
\newblock Designing and interpreting probes with control tasks.
\newblock In Kentaro Inui, Jing Jiang, Vincent Ng, and Xiaojun Wan (eds.),
  \emph{Proceedings of the 2019 Conference on Empirical Methods in Natural
  Language Processing and the 9th International Joint Conference on Natural
  Language Processing (EMNLP-IJCNLP)}, pp.\  2733--2743, Hong Kong, China,
  November 2019. Association for Computational Linguistics.
\newblock \doi{10.18653/v1/D19-1275}.
\newblock URL \url{https://aclanthology.org/D19-1275/}.

\bibitem[Korda \& Mezi{\'c}(2018)Korda and Mezi{\'c}]{korda2018mpc}
Milan Korda and Igor Mezi{\'c}.
\newblock Linear predictors for nonlinear dynamical systems: {K}oopman operator
  meets model predictive control.
\newblock \emph{Automatica}, 93:\penalty0 149--160, 2018.

\bibitem[Lyu et~al.(2023)Lyu, Hu, Siriya, Pu, and Chen]{lyu2023latent}
Xubo Lyu, Hanyang Hu, Seth Siriya, Ye~Pu, and Mo~Chen.
\newblock Task-oriented {K}oopman-based control with contrastive encoder.
\newblock In \emph{Conference on Robot Learning ({CoRL})}, 2023.

\bibitem[Maes et~al.(2026)Maes, Le~Lidec, Scieur, LeCun, and
  Balestriero]{maes2026lewm}
Lucas Maes, Quentin Le~Lidec, Damien Scieur, Yann LeCun, and Randall
  Balestriero.
\newblock {LeWorldModel}: Stable end-to-end joint-embedding predictive
  architecture from pixels, 2026.

\bibitem[Mezi{\'c}(2005)]{mezic2005}
Igor Mezi{\'c}.
\newblock Spectral properties of dynamical systems, model reduction and
  decompositions.
\newblock \emph{Nonlinear Dynamics}, 41\penalty0 (1-3):\penalty0 309--325,
  2005.

\bibitem[Mezi{\'c}(2021)]{mezic2021}
Igor Mezi{\'c}.
\newblock {K}oopman operator, geometry, and learning of dynamical systems.
\newblock \emph{Notices of the American Mathematical Society}, 68\penalty0
  (7):\penalty0 1087--1105, 2021.
\newblock Fully referenced version: arXiv:2010.05377.

\bibitem[Ruiz-Morales et~al.(2025)Ruiz-Morales, Vanoost, Pissoort, and
  Verbeke]{ruizmorales2025}
Pablo Ruiz-Morales, Dries Vanoost, Davy Pissoort, and Mathias Verbeke.
\newblock Koopman invariants as drivers of emergent time-series clustering in
  joint-embedding predictive architectures, 2025.

\bibitem[Silver et~al.(2017)Silver, van Hasselt, Hessel, Schaul, Guez, Harley,
  Dulac-Arnold, Reichert, Rabinowitz, Barreto, and
  Degris]{silver2017predictron}
David Silver, Hado van Hasselt, Matteo Hessel, Tom Schaul, Arthur Guez, Tim
  Harley, Gabriel Dulac-Arnold, David Reichert, Neil Rabinowitz, Andr\'e
  Barreto, and Thomas Degris.
\newblock The predictron: End-to-end learning and planning.
\newblock In \emph{International Conference on Machine Learning}, 2017.

\bibitem[Tassa et~al.(2018)Tassa, Doron, Muldal, Erez, Li, de~Las~Casas,
  Budden, Abdolmaleki, Merel, Lefrancq, Lillicrap, and
  Riedmiller]{tassa2018dmc}
Yuval Tassa, Yotam Doron, Alistair Muldal, Tom Erez, Yazhe Li, Diego
  de~Las~Casas, David Budden, Abbas Abdolmaleki, Josh Merel, Andrew Lefrancq,
  Timothy Lillicrap, and Martin Riedmiller.
\newblock {DeepMind} control suite, 2018.
\newblock arXiv:1801.00690.

\bibitem[Williams et~al.(2015)Williams, Kevrekidis, and
  Rowley]{williams2015edmd}
Matthew~O. Williams, Ioannis~G. Kevrekidis, and Clarence~W. Rowley.
\newblock A data-driven approximation of the {K}oopman operator: Extending
  dynamic mode decomposition.
\newblock \emph{Journal of Nonlinear Science}, 25\penalty0 (6):\penalty0
  1307--1346, 2015.

\end{thebibliography}
\bibliographystyle{rlj}

\clearpage
\appendix

\section{Operator-on-\texorpdfstring{$F$}{F}: full definition}
\label{app:c7}

For an anchor $(s_t,\, a_{t:t+k-1},\, s_{t+k})$ and a functional
$\phi : Z \to \mathbb{R}$ in $F$, the per-anchor operator error is
\[
    e_\phi(t) \;=\; \frac{\bigl|\,\phi(\hat{z}_{t+k}) \,-\, \phi(z_{t+k}')\,\bigr|}{\sigma_\phi}
    , \qquad
    \hat{z}_{t+k} = O_k(z_t,\, a_{t:t+k-1}),
    \quad
    z_{t+k}' = \mathrm{enc}(s_{t+k}),
\]
where $\sigma_\phi$ is the standard deviation of $\phi(z')$ across the anchor
pool. Per-functional aggregation is the root-mean-square over anchors. The
full-$F$ value is the RMS of $\{e_\phi\}_{\phi \in F}$, with $F$ split into
two reported groups:
\begin{itemize}
    \item \textbf{Value slice}: $F = \{r,\, V\}$, where $r$ is the model's
        reward head and $V$ its value head.
    \item \textbf{Full-$F$}: the value slice plus a per-anchor PCA basis on
        the encoded next-state geometry. We use the top-$16$ principal
        directions and whiten by their singular values so each direction
        contributes a unit-variance functional. The PCA is fit on a held-out
        half of the anchor pool; the operator-on-$F$ is evaluated on the other half. This
        prevents the basis from being chosen to minimize operator error on
        the evaluation half.
\end{itemize}
The choice of $k$ matches each model's native planning step: $k = 5$
collect-steps throughout (one frameskip-$5$ predictor call for LeWM, five
open-loop dynamics steps under the recorded actions for TD-MPC2).

\section{Probe protocol}
\label{app:probe}

The probe $\phi_F : Z \to F$ for each observable is a ridge regression
trained on a feature-standardized latent, with the regularizer $\lambda$
chosen by grid search on a held-out sub-split. We report per-observable
$R^2$ alongside each operator-on-$F$ number. For the cross-architecture
comparison in Section~\ref{sec:results-t23} we also restrict $F$ to the
\emph{commonly-well-probed} subset — those observables for which both LeWM
and TD-MPC2 achieve $R^2 \geq 0.7$, giving $n = 7$ observations on
\texttt{cheetah-run}. The restricted-$F$ ratio
$\mathrm{LeWM} / \mathrm{TD\text{-}MPC2}$ is
$0.395 \pm 0.012$ across the three LeWM seeds, compared with
$0.457 \pm 0.011$ on the shared $F$; the gap grows on the restriction, so
probe asymmetry is not the source of the disjoint CIs reported in
Section~\ref{sec:results-t23}.

The unregularized-probe symptom noted in Section~\ref{sec:epistemic} ---
strongly negative $R^2$ on some observables for the $768$-dim mt80-$19$M
latent before regularization --- is what motivates the $\lambda$ grid
search; the chosen $\lambda$ values are recorded per model in the released
configs.

\subsection*{MLP-probe ablation for the cross-architecture comparison}
\label{app:probe-mlp}

The reviewer-relevant question for the cross-architecture comparison in
Section~\ref{sec:results-t23} is whether the LeWM/TD-MPC2 ordering depends
on the probe family. We re-ran the protocol with a $1$-hidden-layer MLP
probe ($64$ ReLU units, weight decay $\alpha$ grid-searched on an $80/20$
sub-split of the train half, identical otherwise to the ridge protocol).
The probe family is the only thing that changed: same compare data, same
$50/50$ train/eval split, same target $F$.

\begin{table}[h]
    \centering
    \caption{MLP-probe ablation of the cross-architecture comparison. The
        absolute operator error on shared $F$ values shift upward under the MLP
        probe (probe family changes the realized observable estimates),
        but the LeWM/TD-MPC2 ordering and the disjoint $95\%$ CIs are
        preserved.}
    \label{tab:t23-mlp}
    \resizebox{\textwidth}{!}{%
    \begin{tabular}{lrrrr}
        \hline
        seed & LeWM op.\ err.\ on shared $F$ & LeWM $95\%$ CI & TD-MPC2 op.\ err.\ on shared $F$ & ratio \\
        \hline
        $1$ & $0.490$ & $[0.476, 0.504]$ & $0.997$ & $0.492$ \\
        $2$ & $0.490$ & $[0.475, 0.505]$ & $0.997$ & $0.492$ \\
        $3$ & $0.500$ & $[0.486, 0.517]$ & $0.997$ & $0.502$ \\
        \hline
        mean $\pm$ SD & $0.494 \pm 0.006$ &  & $0.997$ & $0.495 \pm 0.006$ \\
        \hline
    \end{tabular}}
\end{table}

Compared to the ridge baseline (ratio $0.457 \pm 0.011$,
Table~\ref{tab:t23}), the MLP-probe ratio of $0.495 \pm 0.006$ is slightly
closer to one but the LeWM advantage and disjoint $95\%$ CIs hold. The
cross-architecture gap in Section~\ref{sec:results-t23} is therefore not a
ridge-probe-capacity artifact.

\section{Per-size rows for the TD-MPC2 size sweep}
\label{app:t25}

Table~\ref{tab:t25} reports the five-row source data behind
Figure~\ref{fig:scissors} and the metric-disagreement numbers cited in
Section~\ref{sec:results-t22}.

\begin{table}[h]
    \centering
    \caption{TD-MPC2 mt80 size sweep on \texttt{cheetah-run}. ``reward
        err.'' is the reward-prediction error, ``Bellman res.'' the
        value-equivalence (Bellman) residual, and ``value-only (unnorm.)''
        the unnormalized RMS of $|V(\hat{z}) - V(z')|$, near-identical to
        the Bellman residual across sizes (Spearman $+1.00$). The
        full-$F$ operator-error $95\%$ CI is the anchor-bootstrap
        percentile interval over the $1500$ pooled evaluation anchors
        ($1000$ resamples).}
    \label{tab:t25}
    \resizebox{\textwidth}{!}{%
    \begin{tabular}{lrrrrrr}
        \hline
        size & return & full-$F$ op. & $95\%$ CI & reward err. & Bellman res. & value-only (unnorm.) \\
        \hline
        $1$M   & $13.3$ & $0.91$ & $[0.89, 0.93]$ & $0.042$ & $2.65$ & $2.66$ \\
        $5$M   & $26.1$ & $0.30$ & $[0.28, 0.32]$ & $0.028$ & $1.33$ & $1.33$ \\
        $19$M  & $91.5$ & $0.28$ & $[0.27, 0.29]$ & $0.053$ & $2.78$ & $2.80$ \\
        $48$M  & $43.9$ & $0.36$ & $[0.35, 0.37]$ & $0.031$ & $2.67$ & $2.70$ \\
        $317$M & $\phantom{0}0.9$ & $2.62$ & $[2.43, 2.79]$ & $0.091$ & $6.73$ & $6.78$ \\
        \hline
    \end{tabular}}
\end{table}

Across-size Spearman (full-$F$ operator, return) $= -0.90$
(anchor-bootstrap $95\%$ CI $[-0.90, -0.70]$, $P(\rho < 0) = 1.000$ across
$1000$ resamples). Leave-one-out values: dropping $1$M $\to -0.80$,
$5$M $\to -1.00$, $19$M $\to -0.80$, $48$M $\to -1.00$, $317$M $\to -0.80$,
so the rank order is stable to single-point removal but the $n=5$ sample
remains small. Reward-prediction error has range $[0.028, 0.091]$, so a
partial correlation that conditions on it reproduces the marginal value
(reported $-0.89$); we treat this as consistency, not as an independent
control. Other across-size Spearmans (anchor-bootstrap CIs over the
full-$F$ operator, comparator held at its point estimate): vs.\ Bellman
residual $+0.30$ ($[+0.30, +0.60]$); vs.\ kernel divergence
$+0.10$ ($[+0.10, +0.30]$); for the normalized value-slice $F = \{r, V\}$
vs.\ Bellman residual $+0.10$ ($[+0.10, +0.20]$). Bellman residual vs.\
return: $-0.10$; reward error vs.\ return: $-0.30$.

Under operator-on-$F$'s own per-functional normalization (each channel
divided by the across-anchor spread of its true pushforward), the reward
slice and the value slice $F = \{r, V\}$ track return at Spearman $-0.90$
and $-1.00$ respectively, as strong as the full-$F$ operator ($-0.90$);
their conventional unnormalized forms (raw reward error, Bellman residual)
instead give $-0.30$ and $-0.10$. On this sweep the resolution gap between
operator-on-$F$ and the value-equivalence proxies is thus largely a
normalization effect, and operator-on-$F$'s distinct contribution is to
carry this normalized pushforward comparison onto a shared observable basis
and onto models without reward or value heads
(Section~\ref{sec:results-t23}).

\begin{figure}[h]
    \centering
    \includegraphics[width=0.6\linewidth]{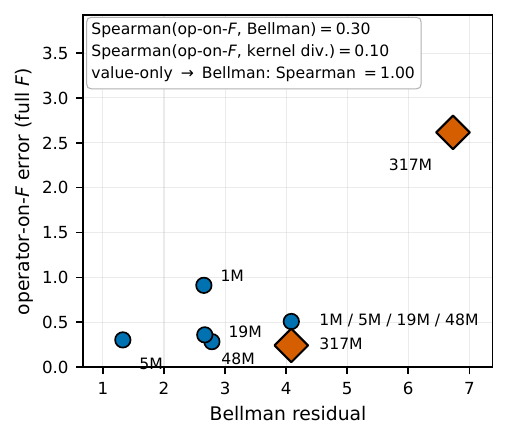}
    \caption{Full-$F$ operator-on-$F$ vs.\ Bellman residual on the
        TD-MPC2 mt80 size sweep. Spearman across the five sizes is
        $+0.30$ for the full-$F$ operator, contrasted with the
        near-identity ($+1.00$) between the unnormalized value-only
        error and the Bellman residual (Table~\ref{tab:t25}). The
        $317$M point is the clearest disagreement.}
    \label{fig:rerank}
\end{figure}

Figure~\ref{fig:rerank} plots the metric-disagreement scatter (full-$F$
operator vs.\ Bellman residual) referenced from
Section~\ref{sec:results-t22}.

\section{Cross-architecture controls}
\label{app:t23}

Table~\ref{tab:t23} reports the three LeWM seeds against the fixed
single-task TD-MPC2 checkpoint. Each cell is generated by re-running the
shared protocol on a checkpoint trained from scratch with a different
random seed.

\begin{table}[h]
    \centering
    \caption{Cross-architecture comparison: per-seed LeWM
        operator error on shared $F$ against the deterministic single-task TD-MPC2
        (operator error on shared $F$) $= 0.840$.}
    \label{tab:t23}
    \resizebox{\textwidth}{!}{%
    \begin{tabular}{lrrrr}
        \hline
        seed & LeWM op.\ err.\ on shared $F$ & LeWM full-$F$ op.\ err.\ & ratio (shared $F$) & restricted-$F$ ratio ($n=7$) \\
        \hline
        $1$ & $0.380$ & $0.396$ & $0.453$ & $0.386$ \\
        $2$ & $0.377$ & $0.417$ & $0.449$ & $0.409$ \\
        $3$ & $0.394$ & $0.408$ & $0.469$ & $0.391$ \\
        \hline
        mean $\pm$ SD & $0.384 \pm 0.009$ & $0.407 \pm 0.010$ & $0.457 \pm 0.011$ & $0.395 \pm 0.012$ \\
        \hline
    \end{tabular}}
\end{table}

As an additional control, we replace the mt80 multitask TD-MPC2 with the
released single-task checkpoint and re-run the comparison; the TD-MPC2 number
is essentially identical ($0.840$ vs.\ $0.833$ on mt80), so the
cross-architecture gap is not a multitask handicap.

\begin{figure}[h]
    \centering
    \includegraphics[width=0.55\linewidth]{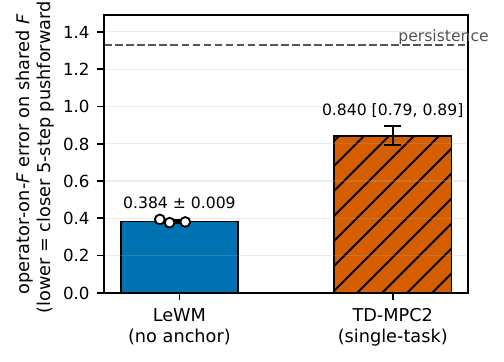}
    \caption{Operator error on shared observation $F$ for LeWM ($3$
        seeds) and single-task TD-MPC2 on \texttt{cheetah-run} at the
        $5$-step horizon. Both models beat the persistence baseline
        (dashed); the $95\%$ CIs are disjoint.}
    \label{fig:crossarch}
\end{figure}

\typeout{get arXiv to do 4 passes: Label(s) may have changed. Rerun}
\end{document}